%% file: main.tex
\documentclass[10pt,twocolumn,letterpaper]{article}

\usepackage{iccv}
\usepackage{times}
\usepackage{epsfig}
\usepackage{graphicx}
\usepackage{amsmath}
\usepackage{amssymb}

\usepackage{makecell} 
\usepackage{multirow}
\usepackage{array}
\newcolumntype{C}[1]{>{\centering\let\newline\\\arraybackslash\hspace{0pt}}m{#1}}
\newcolumntype{L}[1]{>{\raggedright\let\newline\\\arraybackslash\hspace{0pt}}m{#1}}
\newcolumntype{?}{!{\vrule width 1pt}}
\usepackage[table]{xcolor}
\usepackage{graphicx}
\usepackage{amsmath}
\usepackage{amssymb}
\usepackage{booktabs}
\usepackage{times}
\usepackage{epsfig}
\usepackage{nccmath}
\usepackage{pifont}
\usepackage{multirow}
\usepackage{array}
\usepackage{enumerate}
\usepackage{enumitem}
\usepackage{subcaption}
\usepackage{caption}
\usepackage{rotating}
\usepackage[misc]{ifsym}
\usepackage{url}
\usepackage{multirow}
\definecolor{cgreen}{RGB}{34, 139, 34}
\newcommand{\cmark}{\color{cgreen}\ding{51}}%
\newcommand{\xmark}{\color{red}\ding{55}}%

\newcommand{\myb}{\beta}
\newcommand{\myt}{\alpha}
\newcommand{\myg}{t}

\usepackage[pagebackref=true,breaklinks=true,letterpaper=true,colorlinks,bookmarks=false]{hyperref}

\iccvfinalcopy 

\def\iccvPaperID{11448} 

\begin{document}

\title{Animal3D: A Comprehensive Dataset of 3D Animal Pose and Shape}

\author{Jiacong Xu\textsuperscript{1} \qquad Yi Zhang\textsuperscript{1} \qquad Jiawei Peng\textsuperscript{1}\qquad Wufei Ma\textsuperscript{1} \qquad Artur Jesslen\textsuperscript{11} \qquad Pengliang Ji\textsuperscript{3} \\ Qixin Hu\textsuperscript{4} \qquad Jiehua Zhang\textsuperscript{5} \qquad Qihao Liu\textsuperscript{1} \qquad Jiahao Wang\textsuperscript{1} \qquad Wei Ji\textsuperscript{6} \qquad Chen Wang\textsuperscript{7}\\ Xiaoding Yuan\textsuperscript{1} \qquad Prakhar Kaushik\textsuperscript{1} \qquad Guofeng Zhang\textsuperscript{8} \qquad Jie Liu\textsuperscript{9} \qquad Yushan Xie\textsuperscript{2} \\ Yawen Cui\textsuperscript{5}
\qquad Alan Yuille\textsuperscript{1} \qquad Adam Kortylewski\textsuperscript{10,11}  \\\\
\textsuperscript{1}Johns Hopkins University \qquad 
\textsuperscript{2}East China Normal University\qquad 
\textsuperscript{3}Beihang University  \\
\textsuperscript{4}HUST\qquad
\textsuperscript{5}University of Oulu \qquad 
\textsuperscript{6}University of Alberta \qquad 
\textsuperscript{7}Tsinghua University \qquad
\textsuperscript{8}UCLA \\ 
\textsuperscript{9}City University of Hong Kong \qquad 
\textsuperscript{10}Max Planck Institute for Informatics\qquad
\textsuperscript{11}University of Freiburg \qquad 
}

\maketitle

\begin{abstract}
Accurately estimating the 3D pose and shape is an essential step towards understanding animal behavior, and can potentially benefit many downstream applications, such as wildlife conservation.
However, research in this area is held back by the lack of a comprehensive and diverse dataset with high-quality 3D pose and shape annotations.
In this paper, we propose Animal3D, the first comprehensive dataset for mammal animal 3D pose and shape estimation.
Animal3D consists of 3379 images collected from 40 mammal species, high-quality annotations of $26$ keypoints, and importantly the pose and shape parameters of the SMAL \cite{zuffi20173d} model. 
All annotations were labeled and checked manually in a multi-stage process to ensure highest quality results.
Based on the Animal3D dataset, we benchmark representative shape and pose estimation models at: (1) supervised learning from only the Animal3D data, (2) synthetic to real transfer from synthetically generated images, and (3) fine-tuning human pose and shape estimation models. 
Our experimental results demonstrate that predicting the 3D shape and pose of animals across species remains a very challenging task, despite significant advances in human pose estimation. 
Our results further demonstrate that synthetic pre-training is a viable strategy to boost the model performance.
Overall, Animal3D opens new directions for facilitating future research in animal 3D pose and shape estimation, and is publicly available. 
\end{abstract}

\input{tex_files/1_intro}

\input{tex_files/2_related}

\input{tex_files/3_method}
\input{tex_files/4_exp}

\input{tex_files/5_conc}

{\small
\bibliographystyle{ieee_fullname}
\bibliography{egbib}
}

\end{document}

%% file: tex_files/1_intro.tex
\begin{figure*}
    \centering
    \includegraphics[width=\linewidth]{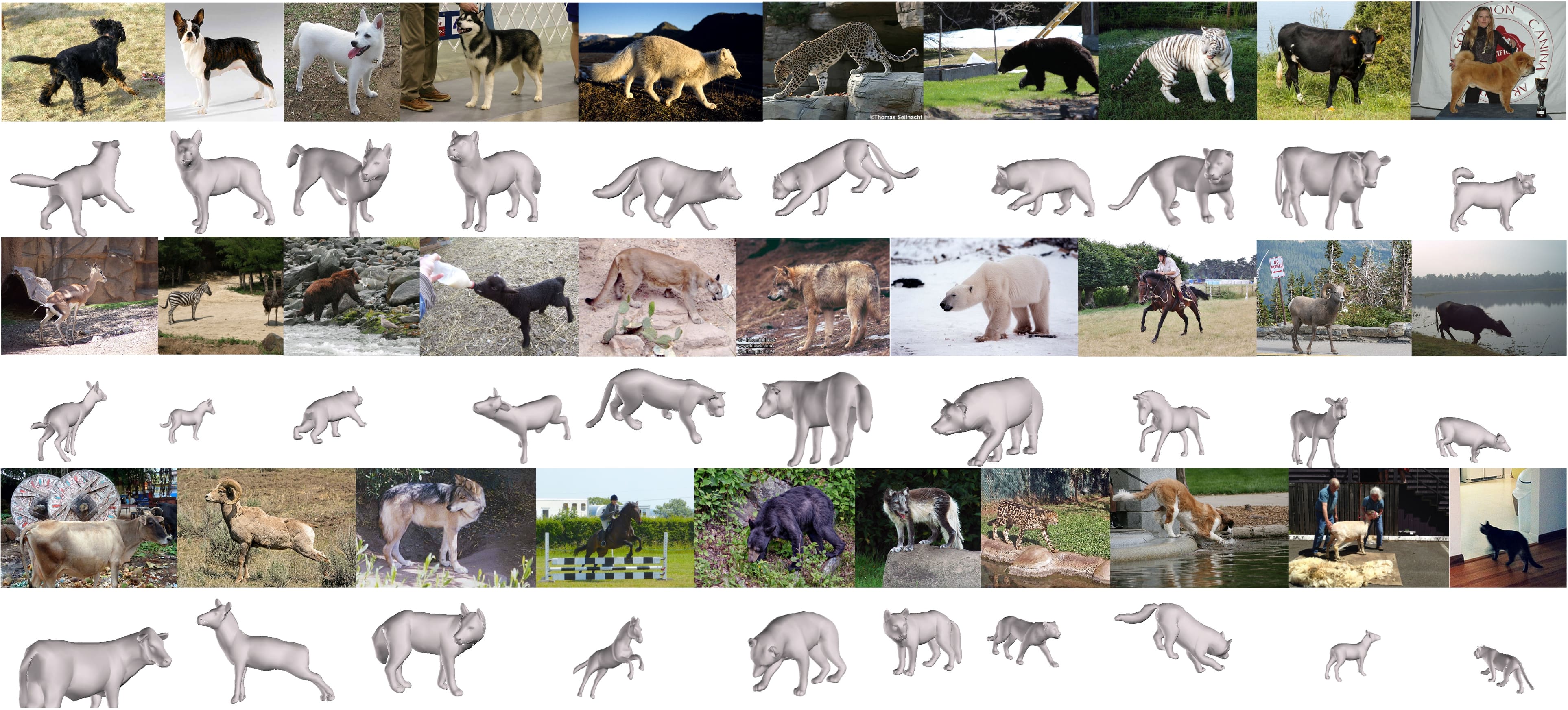}
    \caption{Samples from the proposed Animal3D dataset. Our dataset contains a diverse range of animal species with high-quality annotations of shape and pose parameters using the popular SMAL \cite{zuffi20173d} model.}
    \label{fig:samples}
\end{figure*}

\section{Introduction}
Accurately estimating the 3D pose and shape of animals is a crucial step toward understanding their behavior and has a wide range of applications in fields such as wildlife conservation, animal ecology, and biomechanics.
3D animal pose and shape estimation involves the reconstruction of the 3D structure of an animal from a single 2D image, which is a challenging task due to the complex shapes and poses of animals in the wild.
Previous works in this area have primarily focused on specific animals, such as humans \cite{kanazawa_hmr,kocabas2021pare} or dogs \cite{BARC:2022,yao2022lassie}, which limits the generalization ability of the models to other animals. 
Therefore, there is a need for a diverse dataset of animals to allow for more generalizable and robust models to be developed. 

In this paper, we propose Animal3D, the first benchmark for mammal animal 3D pose and shape estimation.
Animal3D is a comprehensive dataset consisting of $3379$ high-quality images collected from $40$ mammal species. The images were carefully selected from existing datasets, in particular PartImageNet \cite{he2022partimagenet} and COCO \cite{coco}, to ensure that they represent a diverse range of animals, including primates, ungulates, carnivores, and rodents (Figure \ref{fig:samples}).
The diversity of animal species included in the dataset ensures that the models are not limited to specific animals and can be applied to a wide range of species. 
Each image was annotated with $26$ keypoints, which were carefully labeled and checked in a mutli-stage process to ensure high-quality annotations that can be used for further research.
Based on the keypoint annotation and the available segmentation masks in PartImageNet and COCO, we annotate the 3D shape and pose by fitting the SMAL \cite{zuffi20173d} model to the data. SMAL is a widely used model for 3D animal pose and shape estimation and similarly structured as the SMPL \cite{looper_smpl} model, hence supporting wide applicability of our annotations. 

Using the Animal3D dataset, we benchmark representative shape and pose estimation models at three levels: (1) supervised learning for animal pose estimation, (2) synthetic to real transfer from synthetically generated images, and (3) fine-tuning human pose and shape estimation models. 
Based on our experimental results we provide an analysis of the strengths and limitations of each method, which demonstrate the versatility of our benchmark, as well as its challenging nature, since none of the representative approaches achieves a similarly good performance as on the specialized benchmarks they were designed for.

Animal3D is a significant step towards facilitating future research in animal 3D pose and shape estimation. The dataset will allow researchers to advance the understanding of animal behavior and ecology through 3D pose and shape estimation. Additionally, the dataset has the potential to benefit many downstream applications. The models developed using the Animal3D dataset can be applied to a wide range of animals, potentially leading to new discoveries and insights into animal behavior and ecology. Access the data via: \url{https://xujiacong.github.io/Animal3D}

In summary, our main contributions are:
\begin{itemize}
    \item We present Animal3D, the first benchmark for mammal animal 3D pose and shape estimation, with a diverse set of $40$ mammal species, and high-quality annotations of 2D keypoints as well as 3D shape and pose parameters of the SMAL \cite{zuffi20173d} model.
    \item We set up a set of baselines on Animal3D in various settings using state-of-the-art methods which demonstrates the versatility of the dataset.
    \item Our experimental results and in-depth analysis of the strengths and limitations of representative methods demonstrate the challenging nature of our benchmark.
\end{itemize}

%% file: tex_files/2_related.tex
\begin{figure*}
\centering
    \includegraphics[width=\linewidth]{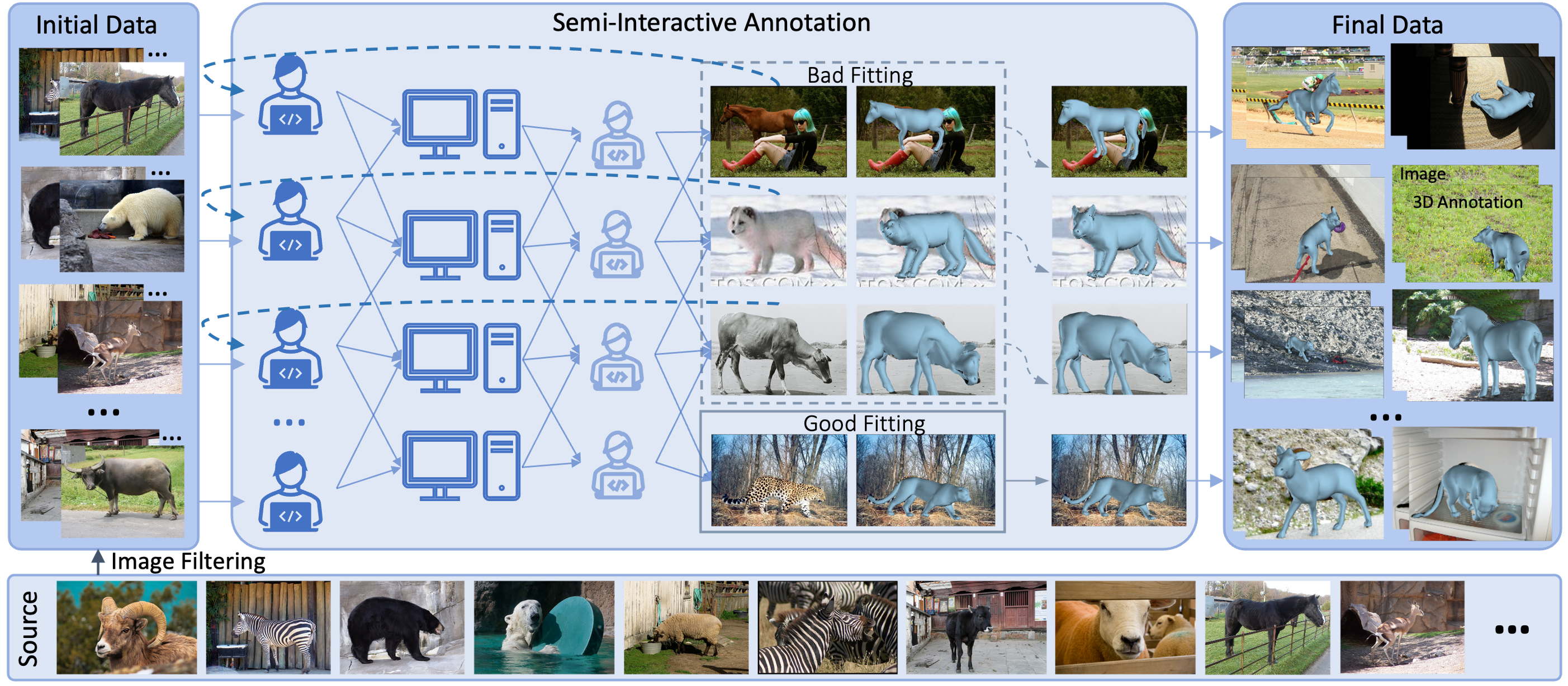}
\caption{Data annotation pipeline for Animal3D. The process consists of three stages: Image Filtering, Semi-Interactive Annotation, and Data Integration. The data is sourced and filtered to obtain an initial set of images.
During the Semi-Interactive Annotation, annotators submitted their annotation to the server to fit the SMAL model and render the results on the images. Then a set of inspectors examined the fitting results and send the bad-fitting images back to the annotator for revision. This process is repeated multiple times. Images that constantly lead to bad-fitting results are removed.}
\label{fig:annotation}
\end{figure*}

\section{Related Work}
Existing animal datasets and the development and challenges of pose estimation methods for both human and animals will be discussed in this section.
\subsection{Animal Pose Estimation}

\textbf{Datasets.} Existing animal datasets provide 2D annotations, such as keypoints, bounding box, or segmentation, on single or multiple species. AP-10K \cite{yu2021ap} is the largest public dataset that contains over 10k images for 54 animal species and 17 keypoints are defined for each animal. Horse-30 \cite{mathis2021pretraining} comprises 8k frames for 30 different horses and 22 keypoints are annotated for each image. AcinoSet \cite{joska2021acinoset} records 119k multi-view frames for Cheetahs and annotates 20 keypoints on 7.6k frames. StanfordExtra \cite{biggs2020wldo} extracts 8.1k images from Stanford dogs and annotates 20 keypoints and segmentation on many dog breeds. Animal Pose Dataset \cite{Cao_2019_ICCV} extends the Poselets dataset by annotating more images from Animal-10 and recently serves as the standard benchmark for 2D animal pose estimation due to its large scale and rich species. Unlike human pose datasets, there is no widely accepted annotation standard for animals since the biological diversity between different animal species is much more significant than among humans. For example, the keypoints of knee in Animal Pose Dataset are defined as middle points on the leg in StanfordExtra, and thereby their positions are slightly different. Nevertheless, 2D keypoint detection is not enough to derive the full geometry information of the objects and there is no known dataset that contains 3D annotation for animals.

\textbf{Methods.} Large amount of data is required for better performance of deep models, but the collection and annotation processes of animal images are much more complex compared with humans. To deal with the data scarcity of animals, researchers came up with many efficient and effective ways. For instance, Cao et al. \cite{Cao_2019_ICCV} feed the models with animal and human data together and employ domain adaptation to align the feature projection space. Mu et al. \cite{mu2020learning} construct a synthetic animal dataset with different textures and poses and utilize the unlabeled images with generated high-score pseudo labels to train the model. To further reduce the domain gap between synthetic and real data, Li and Lee \cite{Li_2021_CVPR} feed the multi-scale information to the domain classifier with gradient reverse layer \cite{ganin2015unsupervised}. These models focus on the 2D keypoint detection task of animals, where benchmarks are available for a fair comparison. 

Nevertheless, the research on 3D pose estimation of animals is proceeding slowly due to the vacancy of available dataset. By mapping the pixels to vertices on template model (CSM), Kulkarni et al. \cite{kulkarni2020articulation} introduce an learning-based approach to optimize the articulation of the template. LASSIE \cite{yao2022lassie} is the first work to recover the shape of articulated object without any template or prior models. The Skinned Multi-Animal Linear (SMAL) model proposed by Zuffi et al. \cite{Zuffi_2017_CVPR} built a parametric way to represent the animal shape and pose based on strong prior and its modeling performance are much better than non-parametric methods. 

Biggs et al. \cite{biggs2019creatures} replace the manual labeling process for keypoints and silhouettes in SMAL by the prediction of pre-trained deep CNNs. By joint optimization on multiple images for the same animal, SMALR \cite{Zuffi_2018_CVPR} recovers more shape and texture details. SMALST \cite{Zuffi:ICCV:2019} directly regresses the shape and pose parameters of SMAL and textures for Zebras, and utilizes the difference between rendered and original images to optimize the neural features. WLDO \cite{biggs2020wldo} and BARC \cite{BARC:2022} are built based on an adapted version of SMAL for dogs and achieve satisfactory 3D recovery performance on various dogs. Since there is no dataset with 3D annotation, aforementioned works can only be evaluated using 2D measures, like 2D keypoints or masks.

\subsection{Human Pose Estimation}
In contrast to animal pose estimation, human pose has been studied extensively in the computer vision literature. 
Regression-based methods~\cite{guler_2019_CVPR,kanazawa_hmr,omran2018nbf,pavlakos2018humanshape,Tan,tung2017self} directly estimate 3D human pose from RGB image using a deep network. Different 3D human pose representations are adopted such as 3D joint locations~\cite{mehta2017monocular,rogez2017lcr}, 3D heatmaps~\cite{Pavlakos17,sun2018integral,zhou2017towards} and parameters of a parametric human body~\cite{kanazawa_hmr,pavlakos2018humanshape,kolotouros2019learning}. 
To model pose ambiguities, e.g. for truncated human images, \cite{biggs2020multibodies,kolotouros2021probabilistic} predict multiple possible poses that have correct 2D projections. Optimization-based methods~\cite{SMPL-X:2019,SPIN:ICCV:2019,bogo2016keep,xiang2019monocular,Xu_2019_ICCV} involve parametric human models like SMPL~\cite{SMPL-X:2019,scape,looper_smpl}, and produce both the 3D human pose and human shape.
The representative method is SMPLify~\cite{bogo2016keep}, which fits the SMPL model to 2D keypoint detections with strong priors. Exploiting more information into the fitting procedure has been investigated, including silhouettes~\cite{lassner_up3d}, multi-view~\cite{huang_mvsmplify}, more expressive shape models~\cite{totalcapture}. \cite{xiang2019monocular} propose to fit 3D part affinity maps to overcome 2D-3D ambiguity.

While there has been a lot of progress in human pose estimation, most methods require large-scale annotated training data and therefore it remains unclear if these approaches can generalize to other animal species.

%% file: tex_files/3_method.tex
\section{Animal3D Dataset}
In this section, we introduce the Animal3D benchmark and discuss the data collection and annotation processes.

Existing annotation methods for 3D human pose estimation datasets \cite{ionescu2013human3,8374605,Marcard_2018_ECCV}, which utilize wearable devices, laser body scanner, multi-camera studio to capture the accurate motion and shape of the humans, cannot be generalized to animals since animals are not as controllable as humans and some are even dangerous. 
To still enable 3D pose and shape annotation of animals, we follow the interactive keypoint annotation idea used in PASCAL3D+ \cite{6836101} for 3D pose annotation of rigid objects.
In particular, we implement a web-based annotation tool that enables fast and accurate keypoint annotations.
We extend it to articulated animals by using the SMAL \cite{Zuffi_2017_CVPR} to fit the 3D animal model to annotated keypoints and segmentation masks. 
Overall, we manually collect and annotate images first, and subsequently conduct three rounds of quality checking and revision process, as illustrated in Figure \ref{fig:annotation}.
Compared to other animal datasets, Animal3D is the first dataset that provides 3D annotation of animals (Table \ref{tab:dataset}).   

\begin{table}[ht]
\centering

\small
\setlength{\tabcolsep}{1mm}
\begin{tabular}{|@{\hspace{2\tabcolsep}}L{18mm}|C{14mm}|C{14mm}|C{12mm}|C{12mm}|}
\hline
                          & \begin{tabular}[c]{@{}c@{}} Animal3D \\ (Ours) \end{tabular} & \begin{tabular}[c]{@{}c@{}} Animal \\ Pose\cite{Cao_2019_ICCV} \end{tabular} & \begin{tabular}[c]{@{}c@{}} Stanford \\ Extra\cite{biggs2020wldo} \end{tabular} & \begin{tabular}[c]{@{}c@{}}AP-10K \\ \cite{yu2021ap} \end{tabular} \\ \hline
Segmentation     & \cmark            & \xmark         & \cmark       & \xmark       \\ \hline
3D Anno.       & \cmark            & \xmark        & \xmark       & \xmark      \\ \hline
\#Species  & 40            & 5         & Dogs        & 54       \\ \hline
\#Keypoints &  $26$             & $20$          & $20$        & $17$       \\ \hline
\#Images &  $3.4K$             & $4K$          & $8.1K$        & $10K$       \\ \hline
\end{tabular}
\caption{Comparison of Animal3D with other animal datasets. Animal3D contains class labels of 40 species, 26 keypoints, and 3D pose and shape parameters from the SMAL model. Totally, there are 5.1k images are carefully annotated in Animal3D, but only 3.4k images are selected after 3-round inspection. The unselected images and annotations will also be published together with Animal3D.}
\label{tab:dataset}
\end{table}

\subsection{Data Collection}

\textbf{Source.} Our aim is to obtain shape and pose parameters by fitting SMAL to images using keypoints and silhouette (foreground segmentation) annotations as described in Section \ref{sec:fitting}. 
To simplify the annotation process, we source the animal images and segmentation masks from existing datasets. 
After investigation of the existing segmentation datasets, we choose PartImageNet \cite{he2022partimagenet} and COCO \cite{lin2014microsoft} as our source datasets since they provide accurate segmentation masks of a diverse set of animals. 

\textbf{Filtering.} Due to the limitation of the PCA shape space of SMAL, some of the animals cannot be represented properly by the SMAL model, such as the elephants and giraffes. Therefore, we remove images belonging to these categories. Additionally, there are a large number of images in which the animals are highly occluded or truncated. This sometimes makes it even challenging for humans to guess the invisible animals' pose or parts correctly. Therefore,we also removed these images from the data.
Finally, we selected a total of 5.1k images of 40 mammal species. Details on the exact animal classes and and image statistics can be found in the supplementary material.

\textbf{Animal class labels.} Unlike PartImageNet where all the images are grouped into ImageNet \cite{deng2009imagenet} categories, COCO does not provide fine-grained class labelling. 
To preserve a detailed category-level annotation, we manually classified the images from COCO into ImageNet categories.

\subsection{Data Annotation}
\begin{figure}
    \centering
   \includegraphics[width=0.99\linewidth]{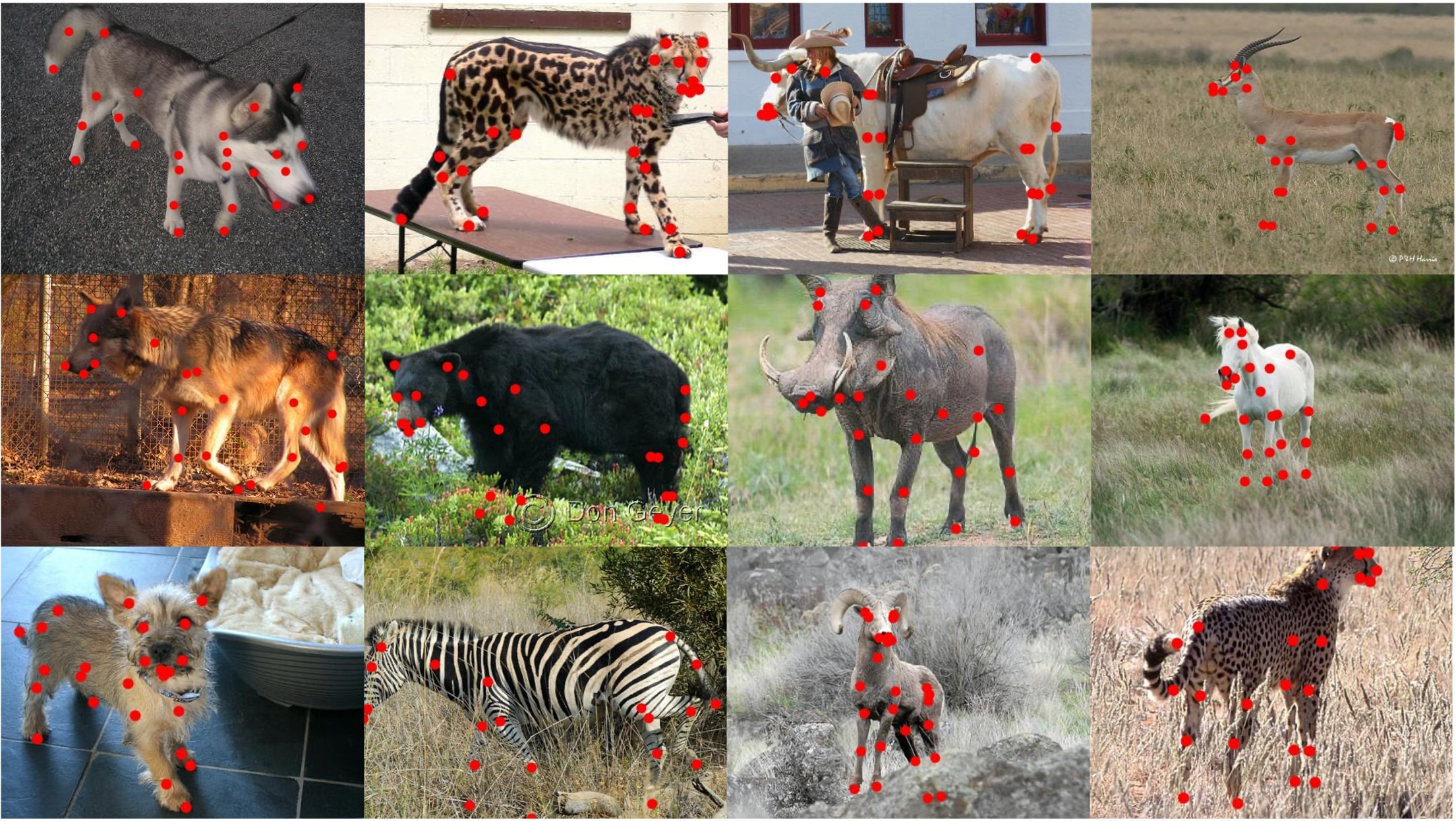}
    \caption{ Visualization of the 26 keypoints that are annotated in the Animal3D model. Other popular 2D datasets only annotate the visible keypoints, while we ask the annotators to guess the location of occluded or truncated body parts based on their annotation experience, which significantly improves the fitting performance of SMAL model.}
    \label{fig:keypoint}
\end{figure}

Since the pose and shape deformation of the SMAL model is highly dependent on the 2D keypoint annotation, we annotate 26 keypoints per animal based on the original keypoint definition of SMAL model (Figure \ref{fig:keypoint}). 
For the keypoint annotation, an interactive is important to guarantee the consistency of the keypoint locations across different images for different animal species, and to make the SMAL fitting results as precise as possible. 
However, the fitting and rendering of the SMAL model cannot be implemented in real-time therefore a fully interactive annotation was not possible.
Instead, we designed an semi-interactive pipeline to make the annotation process as interactive as possible, as described in the following. 

\textbf{Annotators.} Each keypoint annotator were assigned approximately $300$ images belonging to 3-5 similar animal species. They were suggested to start with a few simple cases, where the entire body of the animal is able to be seen clearly, to become familiar with the corresponding animals. For invisible keypoints caused by occlusion, the annotators were asked to guess the positions of the keypoints based on their annotation experience and mark them as invisible. 

\textbf{Inspectors.} The annotation inspectors examine the SMAL fitting results based on the initial annotation to ensure a high annotation quality. They also conduct extra actions to further improve the fitting results. For example, they compare the results with and w/o invisible keypoints, or change hyper-parameters of the SMAL fitting process.

\textbf{Annotation Pipeline.} During the annotation process, the annotators send their annotations to the servers to fit the SMAL model and render the results. 
Subsequently, the inspectors assess the annotation quality, filtering out the good-fitting examples and from the bad-fittings. 
If the bad fitting is caused by the keypoint annotation, the inspectors send the images back to the annotators for revision, and provide feedback on where the annotation can be imroved.
If the bad fitting is caused by a broken segmentation mask from slightly occluded or truncated objects, the inspectors gradually decrease the weights of silhouette error in the objective function to reduce the effect of mask, which in these cases typically improves the fitting results significantly. 
This annotate-then-examine process will proceed for three rounds.
After the final round, the remaining bad-fitting images were discarded.

\subsection{Fitting Animals to Images}\label{sec:fitting}
The SMAL model $M(\myb, \myt, \myg)$ is a function of shape $\myb$, pose $\myt$, and  translation $\myg$. 
We fit the model to images by optimizing the model parameters guided by a combination of 2D keypoints and 2D silhouettes, as proposed in \cite{zuffi20173d}, with minor modifications. 
In the following, we provide a concise description of the fitting process, more details can be found in the original work. 
We denote $P(\textbf{v}_i)$ as the perspective projection of the the $i$'th mesh vertex into the image plane.
Moreover, $P(M) = S$ is the projected mesh silhouette. 
To fit SMAL to an image, we optimize the model parameters $\Pi=\{\myb, \myt, \myg\}$ over a joint loss function that is composed of the reprojection error of the keypoint, the silhouette reprojection error, and a shape prior
\begin{equation}
\mathcal{L}_{total}(\Pi,M)= \mathcal{L}_{kp}(\Pi,M) + \mathcal{L}_{silh}(\Pi,M) + \mathcal{L}_{shape}(\myb).
\label{eq:fit}    
\end{equation}
The losses are weighted to be approx. in the same range.

\textbf{Keypoint loss.} Each keypoint on the SMAL model corresponds to a subset of the mesh vertices. We denote this set of keypoint vertices as $\textbf{v}^i_j \in V_i$, and optimize their projection to match the corresponding annotated keypoint $t_i$:
\begin{equation}
  \label{eq:kp}
\mathcal{L}_{kp}(\Pi,M) = \sum_{i=1}^{26} \rho(|| \sum_{j=1}^{N^i} P(\textbf{v}^i_j)/N^i - t_i||_2),
\end{equation}
where $\rho$ is the Geman-McClure robust error function \cite{geman1987statistical} to reduce the negative effects of difficult to fit annotations.

\textbf{Silhouette loss.} The silhouette is optimized using a bi-directional distance:
 \begin{equation}
  \label{eq:silh}
  \mathcal{L}_{silh}(\Pi,M) ={} \sum_{x \in S}\mathcal{D}_{\bar{S}}(x) +
  \sum_{x\in S} \rho(\min_{\hat{x} \in S} ||x - \hat{x} ||_2),
\end{equation}
where $\bar{S}$ is the ground truth silhouette and $\mathcal{D}_{\bar{S}}$ is its distance transform. The weight for the first term are manually adjusted to reduce the effect of occlusion and truncation.

\noindent{\bf Shape prior.}
We regularize the shape parameters $\myb$ using a shape loss $\mathcal{L}_{shape}(\myb)$ using the PCA prior distribution. In paricular, the loss is defined to be the squared Mahalanobis distance defined using the PCA eigenvalues.

\subsection{Data Summary}
Building on previous datasets and our additional annotation, the Animal3D dataset presents a comprehensive set of annotations for each image, including detailed Imagenet class labels, segmentation masks, 26 keypoints, and SMAL parameters for shape, pose and translation. 
Therefore, Animal3D can serve as a benchmark for a number of tasks as well as multi-tasking, 3D reconstruction or synthetic to real domain adaptation. We believe that our dataset will enable significant advances in all of these research areas, due to its high-quality annotation and scale.

%% file: tex_files/4_exp.tex
\begin{table*}
\centering
\begin{tabular}{@{}l@{}c@{}c@{}c@{}c@{}c@{}c@{}c@{}c@{}c@{}}
\toprule
\multirow{2}{*}{Method} & \multicolumn{3}{c}{Supervised} &  \multicolumn{3}{c}{Synthetic to Real} &  \multicolumn{3}{c}{Human pre-trained} \\ \cmidrule{2-4}  \cmidrule{5-7}  \cmidrule{8-10}
 & \multicolumn{1}{c}{PA-MPJPE$\downarrow$} & \multicolumn{1}{c}{S-MPJPE$\downarrow$} & PCK$\uparrow$ & \multicolumn{1}{c}{PA-MPJPE$\downarrow$} & \multicolumn{1}{c}{S-MPJPE$\downarrow$} & PCK$\uparrow$ & \multicolumn{1}{c}{PA-MPJPE$\downarrow$} & \multicolumn{1}{c}{S-MPJPE$\downarrow$} & PCK$\uparrow$ \\ \midrule
HMR~\cite{kanazawa_hmr} & 140.7 & 496.2 & 59.3 & 124.8 & 497.7 & 63.1 & 132.2 & 488.0 & 60.6 \\ 
PARE~\cite{kocabas2021pare} & 134.8 & 443.9 & 79.1 & 127.2 & 392.3  & 83.7  & 130.7  & 374.9  & 85.6  \\ 
WLDO~\cite{biggs2020wldo} & 128.8  &  502.1 & 60.1  &  123.9 &  484.0 & 65.1  & - & - & - \\ 
 \bottomrule
\end{tabular}
\caption{\label{tab:results} 3D pose and shape estimation results on the Animal3D dataset. We evaluate three representative baseline models, HMR, PARE and WLDO, in three settings: (1) Supervised on Animal3D data only, (2) Pre-training on synthetic data and fine-tuning on Animal3D, and (3) Pre-training on Human Pose Estimation datasets and fine-tuning on Animal3D. While pre-training improves results for all models, the final results are lower compared to object specific benchmarks for humans and dogs, hence indicating the difficulty of estimating 3D animal pose across species. }
\end{table*}

\section{Experiments}
In this section, we benchmark representative shape
and pose estimation models on Animal3D in three settings: (1) supervised learning (Section \ref{sec:exp:pose}), (2) synthetic to real transfer from synthetically generated images (Section \ref{sec:exp:synth}), and (3) fine-tuning pre-trained human pose and shape estimation models (Section \ref{sec:exp:human}). 

\textbf{Baselines.} 
As we present the first comprehensive dataset of 3D shape and pose annotations for animals, there are no baselines that were explicitly designed for animal pose estimation in such a diverse setting. 
Nevertheless, strong representative baselines exist for human pose estimation and for 3D pose estimation of specific animal classes, which we adapt to the Animal3D dataset. 
We chose HMR \cite{kanazawa_hmr} and PARE \cite{kocabas2021pare} as competitive and robust baselines for human pose estimation.
Moreover, we selected WLDO \cite{biggs2020wldo} as a baseline that was specifically designed for animals, although only for dogs. 

\textbf{Evaluation metrics.} We report scale-aligned mean per joint position error (S-MPJPE) and Procrustes-aligned mean per joint position error (PA-MPJPE) in mm as the main evaluation metrics, where the latter is the former plus rotational alignment. We do not use the popular per joint position error (MPJPE) in 3D human pose estimation since the scale of animals can vary a lot. 
We also report the 2D Percentage of Correct Keypoints (PCK) with threshold defined by half of the head-to-tail length to measure how well the prediction aligns with the 2D image.

\textbf{Model \& Data Preparation.}
We split the Animal3D dataset into $3059$ training and $320$ test images, by randomly sampling from the full dataset.

For training HMR \cite{kanazawa_hmr}, we remove its discrimination loss and keep only the 2D and 3D supervision, since there is no fake and real pose parameter available. PARE \cite{kocabas2021pare} requires the part grouping of the model vertices to render the ground-truth of 2D part segmentation. To obtain these labels, we manually segment all the vertices on SMAL into the $7$ object parts defined in PartImageNet \cite{he2022partimagenet}: Head, Torso, Tail, and the four legs. 
To train WLDO \cite{biggs2020wldo}, we remove the EM process that was designed to deal with different dog families, and we supply WLDO with direct supervision of shape and pose parameters for fair comparison with the human models. 

\textbf{Training setup.}
To be consistent with existing training pipelines for human pose estimation models, the animals are cropped from the image based on their bounding box and resized to $224\times 224$. Random rotation and flipping are implemented for data augmentation. The batch size for all the experiments are set to be 128 and we trained the models on 2 RTX3090 GPUs using synchronized batch normalization. For experiments with synthetic data, we pre-train the models for 100 epochs. For training on real data, all the models are trained for 1000 epochs (around 24k iterations).

\subsection{Supervised Animal Shape and Pose Regression}\label{sec:exp:pose}
The left part of Table \ref{tab:results} shows the results of training HMR, PARE and WLDO in a supervised manner only from images of the Animal3D dataset. 
The ranking among the methods is as expected. HMR, which is an older method, performs worse compared to recently developed PARE model. Although the performance gap in PA-MPJPE is smaller compared to the respective results on human pose estimation datasets.
WLDO performs best in terms of PA-MPJPE, hence suggesting that it predicts the articulation of the animals most accurately. However it does not perform particularly well at predicting the rigid 3D body pose, hence it achieves the worst results in terms of S-MPJPE.
Notably, we observe that the prediction accuracy of all baseline models is significantly lower compared to their performance on the original domains that they were initially designed for. Hence, pointing out that the \textbf{3D animal pose estimation problem remains an important open research problem}.
We believe that this is performance gap is caused mainly by two main factors. The lack of large-scale annotated datasets, and the architectural design of the baseline methods for a particular object class, i.e. humans and dogs.
In, the following we aim to address the data problem using synthetic data and pre-training on large scale human data.

\begin{figure*}
\centering
    \includegraphics[width=\textwidth]{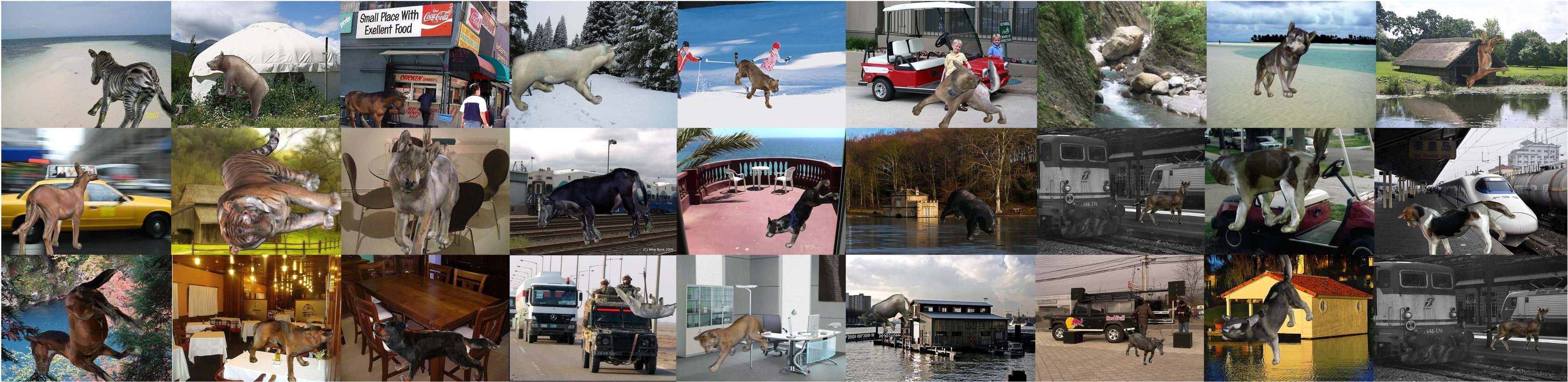}
\caption{Example images from our synthetic dataset that is used for pre-training the animal pose estimation baselines. We simulate all species from the Animal3D dataset using the SMALR model in varying poses, shapes, and background images.}
\label{fig:syn_imgs}
\end{figure*}

\subsection{From Synthetic to Real}\label{sec:exp:synth}
For human pose estimation, a larger amount of high-quality data usually will lead to better regression performance. Nevertheless, the annotation process for Animal3D is too complex to make it a larger dataset, so we are seeking more convenient way to generate more usable data. Inspired by \cite{Mu_2020_CVPR}, which utilize rendered images from CAD model of animals to boost the model performance on 2D tasks, we synthesize 45k images using SMALR \cite{Zuffi_2018_CVPR} and select 40k for training and 5k for testing models, respectively, before fine-tuning them on Animal3D.

For each class in our dataset, several unoccluded and non-truncated images are selected to fit the SMALR model for the basic shape and texture. Then, we calculate the mean and covariance of the shape and pose parameters in the training set and sample the parameters from multi-variate Gaussian to mimic the realistic shape and pose for corresponding animal category. The background images are randomly selected from ImageNet \cite{deng2009imagenet} and consist of indoor and outdoor scenarios. Here are some examples for the synthetic data in Figure \ref{fig:syn_imgs}. Note that, the generation process of the synthetic data is based on the prior information obtained from Animal3D dataset, so its searching space can be enlarged by moderate the covariance. The center part of Table \ref{tab:results} shows that all methods benefit from synthetic pre-training. PARE benefits the most on average across metrics, outperforms HMR and WLDO significantly in terms of S-MPJPE and PCK. and thereby shows the potential of synthetic pre-training for animal pose estimation.

\subsection{From Human to Animal}\label{sec:exp:human}

A common approach to training deep networks in a data efficient manner, is to initialize with models that are pre-trained on large datasets in related tasks. 
We study the effect of using pre-trained human pose estimation models as initialization 
to train an animal pose estimator.
Both HMR and PARE have been trained on large-scale human data including \texttt{Human3.6M}~\cite{ionescu2013human3}, \texttt{MPI-INF-3DHP}~\cite{mehta2017monocular}, and \texttt{COCO}~\cite{lin2014microsoft} datasets, and we use the publicly available models to fine-tune them on Animal3D.

The right part of Table \ref{tab:results} shows that both models outperform their non-pre-trained counterparts. Interestingly, the performance gap between HMR and PARE model in terms of S-MPJPE and PCK increases due to human pre-training. However, compared to the pre-training on synthetic data, which is much more easy to achieve, pre-training on real human data does not show a benefit.

\subsection{Qualitative Results}
\begin{figure*}[h]
\centering
    \includegraphics[width=\linewidth]{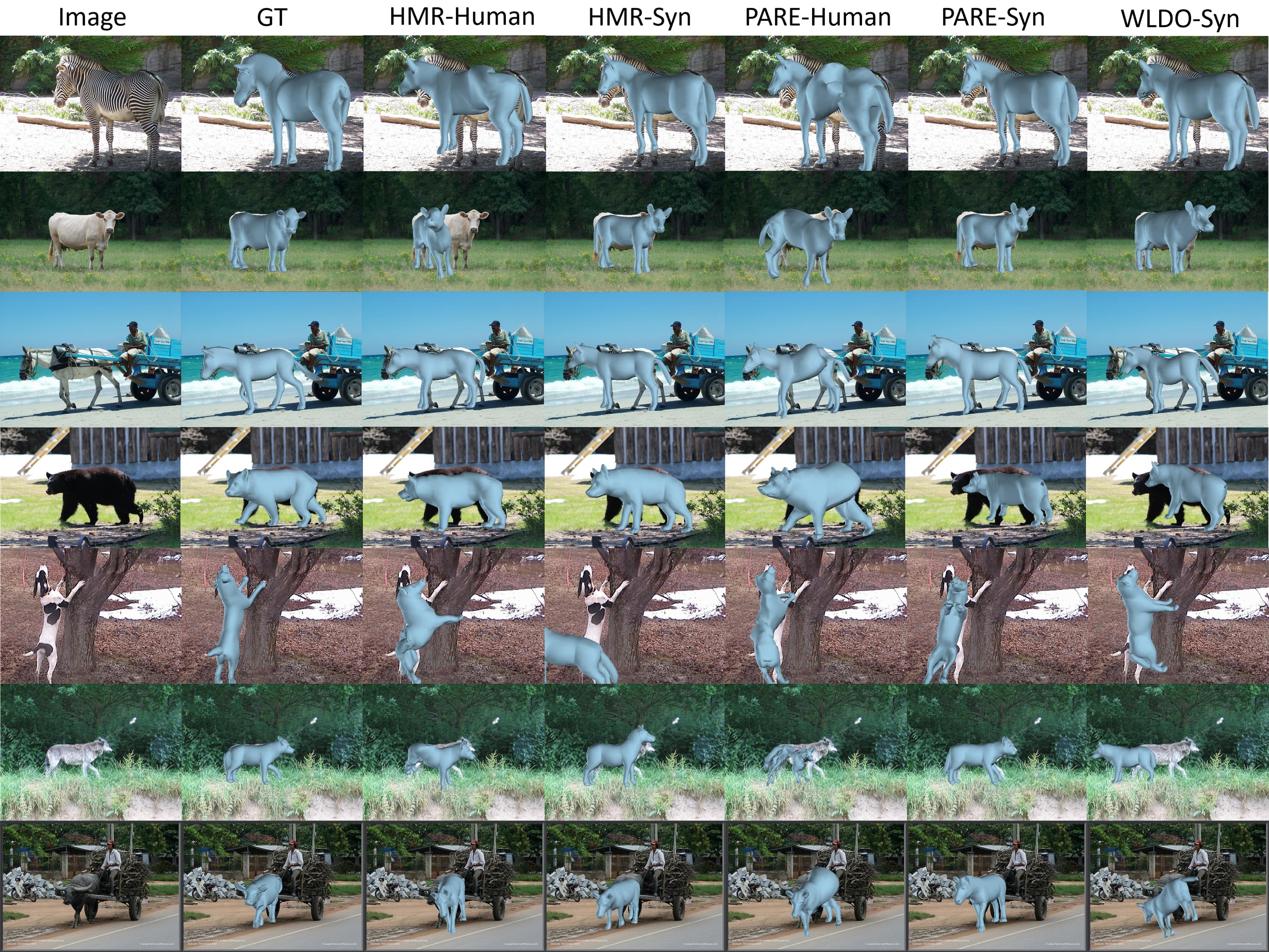}
\caption{Visualization of regression performance of human and animal pose estimation models. The columns from left to right refer to the input image, the groundtruth from Animal3D, regression results for HMR, HMR pretrained by synthetic and human data, PARE and WLDO pretrained by synthetic data, respectively. }
\label{fig:quality}
\end{figure*}

Figure \ref{fig:quality} illustrates qualitative regression results of several models that we have tested in Table \ref{tab:results}.
We observe that human-pretrained models always fail to recover the shape information of the animals and generate some unrealistic shapes. On the contrary, the models pretrained by synthetic data regress the shape parameters better. We argue that the model is able to learn strong shape prior from synthetic data, while it will focus more on the pose information for human data since the shape diversity between humans are much smaller than among animals. Also, the domain gap between human and animals and different feature projection space of SMPL \cite{looper_smpl} and SMAL will hinder the generalization of the models. Even so, in most cases the shape parameters are approximately correctly estimated, i.e. often the correct animal species is predicted. However, the alignment of the legs and the gaze direction are often incorrect. 
Overall, PARE demonstrates also qualitatively that its predicitons have the best quality, mostly because they align better to the image, which can also be observed from its high PCK.

\subsection{Discussion}

Based on our results, we observe that PARE is a promising model for 3D animal pose estimation, as it achieves the best performance among the representative baselines.
Its high performance is very likely caused by its advanced architecture that uses additional supervision in terms of part segmentations. 
In terms of scaling deep learning to animal 3D pose estimation, our results show that pre-training on large-scale synthetic data is a promising direction forward. 
Nevertheless, we observe that none of the baselines achieves a satisfying performance compared to the results obtained on the specific domains that the baselines were originally designed for.
Hence, our experimental results demonstrate that predicting the 3D shape and pose of animals across species remains a very challenging task, despite significant advances in human pose estimation and animal pose estimation for specific species.

%% file: tex_files/5_conc.tex
\section{Conclusion}

Animal3D is unique and diverse in that it includes 3D annotations for a large number of animal species, making it the first benchmark for mammal animal 3D pose and shape estimation. 
The comprehensive nature of Animal3D, in terms of diversity of animals and annotations for multiple vision tasks (keypoints, 3D SMAL parameters, segmentation) provides a foundation for the development of more robust and generalizable models for animal 3D pose and shape estimation. Animal3D and the implementation of the baselines will be publicly available, and we encourage researchers to use it for further research or applications.

The results of our experiments demonstrate that Animal3D is a valuable resource for improving animal pose and shape estimation models. We show that our dataset can be used to benchmark supervised learning for animal pose estimation, synthetic to real transfer, and fine-tuning human pose and shape estimation models. 
We observe that existing methods for human pose estimation, achieve competitive results at animal pose estimation, when pretrained on synthetic data. 
However, the prediction performance is significantly lower compared to their accuracy on human-specific benchmarks. 
These results highlight that the 3D animal pose estimation task remains an important open research problem.
These experiments will provide a strong foundation for future research in this area, which will benefit both scientific understanding and conservation efforts.

\textbf{Acknowledgements.} Adam Kortylewski acknowledges support via his Emmy Noether Research Group funded by the German Science Foundation (DFG) under Grant No. 468670075. Alan Yuille acknowledges support from Army Research Laboratory award W911NF2320008 and Office of Naval Research N00014-21-1-2812.